\makeatletter\def\graphicscache@inhibit{true}\makeatother

\documentclass[letterpaper, 10 pt, conference]{ieeeconf}  %

\IEEEoverridecommandlockouts                              %

\usepackage[utf8]{inputenc}

\usepackage[hidelinks]{hyperref}
\usepackage[style=ieee,hyperref,natbib=true,backend=bibtex,firstinits,doi=false,%
     mincitenames=1,maxcitenames=2,maxbibnames=99,sorting=none,terseinits=false,hyperref=true]{biblatex}
\bibliography{references.bib}
\renewbibmacro*{bbx:savehash}{}%
\defbibheading{bibliography}[\bibname]{\section*{References}}

\usepackage{url}
\usepackage{graphicx}
\usepackage[usenames,dvipsnames]{xcolor}
\usepackage[draft]{fixme}
\fxsetup{theme=color}

\usepackage{amsmath}
\usepackage{amssymb}
\usepackage{textcomp}
\usepackage{siunitx}

\usepackage{booktabs}
\usepackage{threeparttable}
\usepackage{multirow}

\usepackage[capitalize]{cleveref}

\usepackage{tikz}
\usetikzlibrary{arrows}
\usetikzlibrary{positioning,calc}
\usetikzlibrary{decorations.pathreplacing}
\usetikzlibrary{decorations.markings}
\usetikzlibrary{fit}
\usetikzlibrary{shapes.callouts}
\usetikzlibrary{shapes.geometric}
\usetikzlibrary{matrix}
\usepackage{setspace}

\usepackage{pgfplots}
\pgfplotsset{compat=1.9}
\usepgfplotslibrary{groupplots}
\usepgfplotslibrary{units}

\usepackage{balance}

\usepackage{blindtext}

\usepackage{ifthen}

\usepackage[export]{adjustbox}

\IfFileExists{graphicscache.sty}{\usepackage{graphicscache}}

\usepackage{placeins}
\usepackage{dblfloatfix}

\title{\LARGE \bf
Stillleben: Realistic Scene Synthesis for Deep Learning in Robotics
}

\author{Max Schwarz$^{*}$ and Sven Behnke%
\thanks{$^{*}$All authors are with the Autonomous Intelligent Systems group of University of Bonn, Germany; {\tt schwarz@ais.uni-bonn.de}}%
}

\begin{document}

\maketitle

\begin{abstract}

Training data is the key ingredient for deep learning approaches, but difficult
to obtain for the specialized domains often encountered in robotics.
We describe a synthesis pipeline capable of producing training data for
cluttered scene perception tasks such as semantic segmentation, object detection,
and correspondence or pose estimation.
Our approach arranges object meshes in physically realistic, dense scenes using
physics simulation. The arranged scenes are rendered using high-quality
rasterization with randomized appearance and material parameters.
Noise and other transformations introduced by the camera sensors are simulated.
Our pipeline can be run online during training of a deep neural network, yielding
applications in life-long learning and in iterative render-and-compare approaches.
We demonstrate the usability by learning semantic segmentation on the challenging
YCB-Video dataset without actually using any training frames, where our method
achieves performance comparable to a conventionally trained model.
Additionally, we show successful application in a real-world regrasping system.

\end{abstract}

\section{Introduction}

While the rise of deep learning for computer vision tasks has brought new
capabilities to roboticists, such as robust scene segmentation, pose estimation,
grasp planning, and many more, one of the key problems is that deep learning
methods usually require large-scale training datasets. This is less of
a problem for the computer vision community, where researchers can work on
the available public datasets, but roboticists face a key restriction:
Their methods usually need to be deployed in a particular domain, which is
often not covered by the available large-scale datasets.
Capturing a custom training dataset just for one specific purpose is often
infeasible, because it involves careful planning, scene building, capturing,
and usually manual annotation.

There are techniques for reducing the amount of required training data. Transfer
learning, usually in the form of fine-tuning, where a network fully trained on
a generic, large dataset is further trained on a smaller domain-specific
dataset is the preferred method in these situations, as e.g. evident by the
winning approaches at the Amazon Robotics Challenge (ARC)
2017~\citep{morrison2018cartman,schwarz2018fast}.
Here, the dominant methods performed their fine-tuning on synthetic datasets
generated from monocular object images. While picking novel items after roughly
30\,min of capture/training time is an impressive feat, these methods
are limited by the 2D composition of their synthetic scenes. The resulting
arrangements are often not physically realistic and fail to model key effects,
such as correct lighting, shadows, and contact interactions.
Furthermore, the resulting images are only annotated with ground truth
segmentation, prohibiting the training of more abstract tasks like
pose or correspondence estimation.

Still, bolstered by the success of synthetic data generation during the ARC, we
want to ask the question: Are large-scale image datasets still necessary for
robotic perception?

To address the problems of the mentioned 2D synthesis methods, we extend
the idea to 3D. We propose a scene synthesis pipeline consisting of physically
realistic placement, realistic 3D rendering, modeling of camera effects
and automatic ground truth data generation for a variety of tasks.
Our system focuses on highly cluttered arrangements of objects on a planar
supporting surface, although it can be easily adapted to various support or
container shapes. A physics engine is used to place the objects on the support
surface. The scene is then rendered using standard GPU rasterization techniques.

\begin{figure}
 \centering
 \newlength{\meshimgw}\setlength{\meshimgw}{.6cm}
 \begin{tikzpicture}[
     font={\sffamily\scriptsize},
     n/.style={rounded corners, draw=black, align=center},
     proc/.style={n,minimum width=2.2cm, fill=yellow!20},
     ]

     \node[n, align=left, anchor=north] at (0.4,1.8) (meshes) {
		Object meshes:\\[.1cm]
		$\bullet$ Mesh datasets \\
		$\bullet$ 3D capture \\
		$\bullet$ Online DBs \\[0.3cm]

		\includegraphics[width=\meshimgw]{images/meshes/002_master_chef_can.jpg}%
		\includegraphics[width=\meshimgw]{images/meshes/003_cracker_box.jpg}%
		\includegraphics[width=\meshimgw]{images/meshes/004_sugar_box.jpg}\\

		\includegraphics[width=\meshimgw]{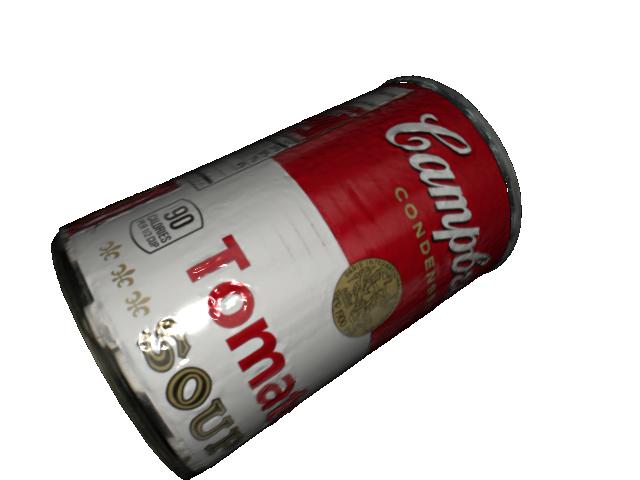}%
		\includegraphics[width=\meshimgw]{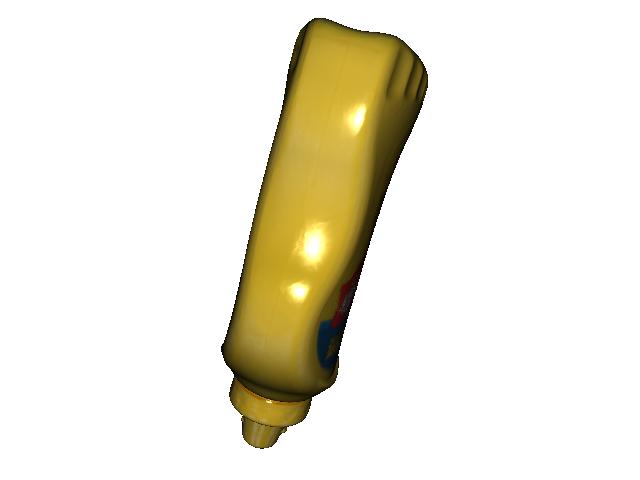}%
		\includegraphics[width=\meshimgw]{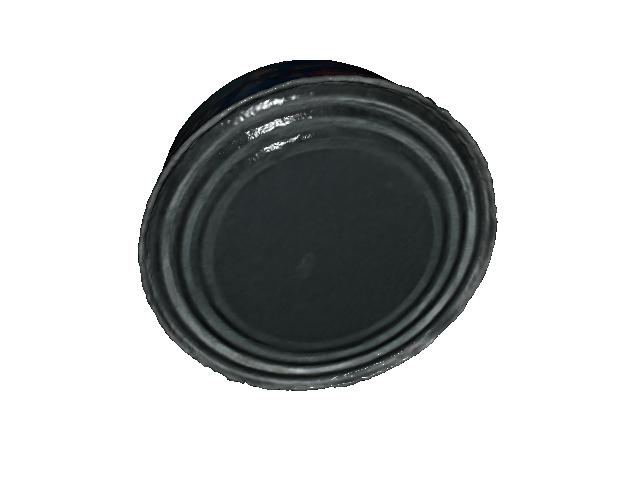}\\

		\includegraphics[width=\meshimgw]{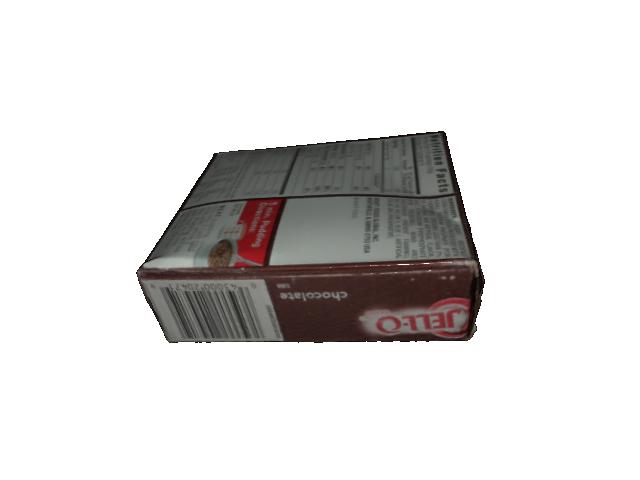}%
		\includegraphics[width=\meshimgw]{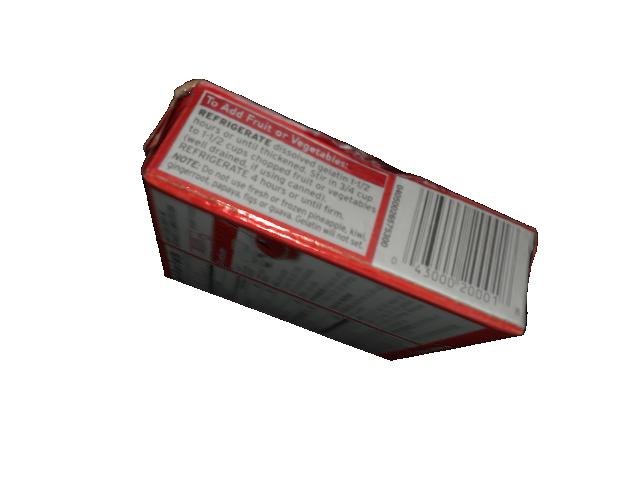}%
		\includegraphics[width=\meshimgw]{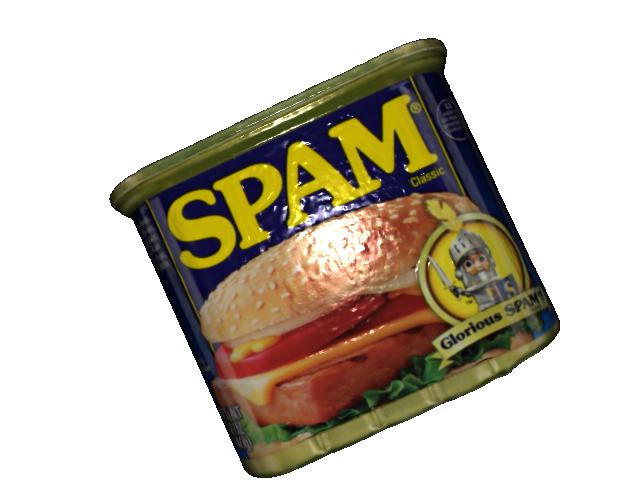}\\

		\includegraphics[width=\meshimgw]{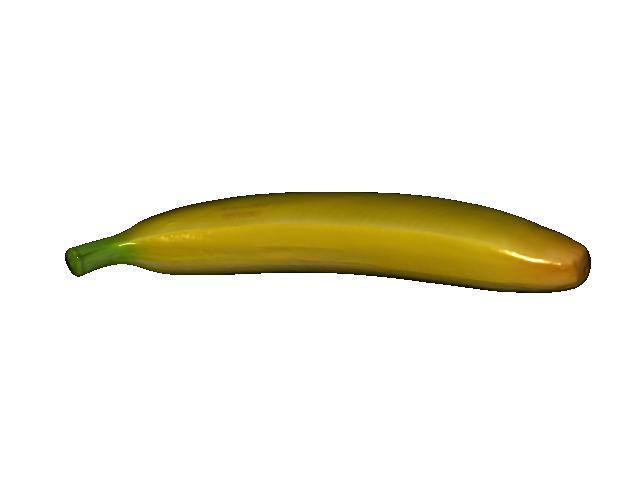}%
		\includegraphics[width=\meshimgw]{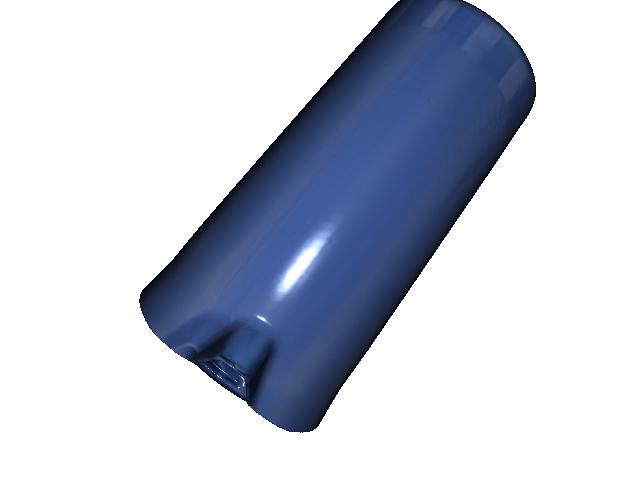}%
		\includegraphics[width=\meshimgw]{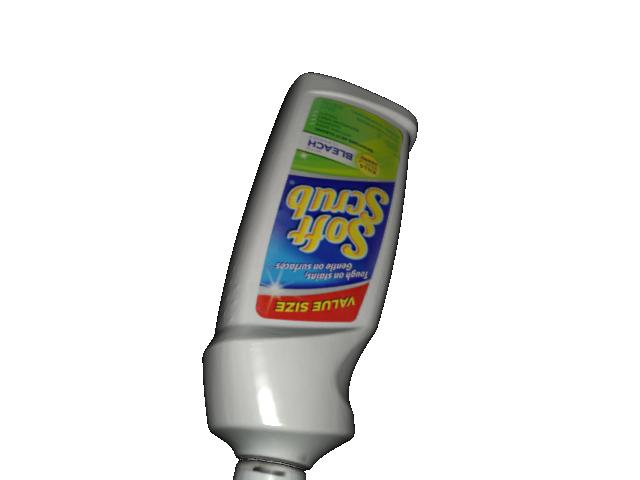}
		};

     \node[proc,anchor=north] (arrangement) at (3,1.8) {1) Arrangement\\Engine (CPU)};

     \node[proc] (rendering) at (3,0) {2) Rendering\\(GPU)};

     \node[proc] (cam) at (3,-1.4) {3) Camera model\\(GPU)};

     \node[n] (result) at (5.8, 0) {Training Frame (GPU)\\[.1cm]
     \includegraphics[frame,width=2cm,clip,trim=50 100 50 0]{images/generator/frame1_rgb.png}\\
     \includegraphics[frame,width=2cm,clip,trim=50 100 50 0]{images/generator/frame1_label.png}};

     \node[proc,rotate=90] (training) at (7.6, 0) {Usual training pipeline (GPU)};

     \draw[-latex] (meshes.east|-arrangement) -- (arrangement.west);

     \draw[-latex] (arrangement) -- (rendering) node[midway,right,align=left] {Arranged\\scene};

     \draw[-latex] (rendering) -- (cam) node[midway,right,align=left] {Rendered\\scene};

     \draw[-latex] (cam) -- (result.west|-cam);

     \draw[-latex] (result) -- (training);

   \end{tikzpicture}\vspace{-2mm}
   \caption{Architecture of our online scene synthesis library.
   Meshes from various sources are arranged in realistic configurations by
   the Arrangement Engine. The render module generates high-quality renders,
   which are then imbued with typical noise and transformations caused by
   commodity cameras. Finally, a usual training pipeline can follow.}
   \label{fig:architecture}
\end{figure}

In short, our contributions include:
\begin{enumerate}
 \item An online, highly efficient scene synthesis pipeline with automatic ground
   truth annotation for training deep learning models,
 \item application of said pipeline to the task of scene segmentation on the
   challenging YCB-Video dataset without actually using any training frames,
   where it reaches comparable performance to a conventionally-trained model, and
 \item application in a real-world robotic system, where it is successfully used to train
   a combined segmentation and pose estimation network.
\end{enumerate}

Our scene synthesis library named \textit{stillleben}, after the German term
for still live paintings, contains bindings to the popular PyTorch deep learning
framework and is available as open source\footnote{\url{https://github.com/AIS-Bonn/stillleben}}.
It will thus hopefully become a helpful tool for robotics researchers.

\section{Related Work}

One of the inspiring works for us is the introduction of \textit{Domain Randomization}
by \citet{tobin2017domain}, who demonstrated learning of an object detection
task on entirely synthetic data, with later execution on a real robotic system.
The key insight by the authors is that the reality gap between synthetic and
real data can be bridged by randomizing the parameters of the simulation generating
the synthetic scenes. With enough ``spread'' of the parameters, situations
close to the real domain will be generated and learned.
In contrast to their method, which is limited to non-cluttered arrangements of
simple shapes in a fixed 2D scene, our method offers a flexible way to
simulate dense, cluttered arrangements in 3D, while modeling and randomizing
more visual effects.
\Citet{andrychowicz2018learning} demonstrate applicability of the domain
randomization idea to a manipulation task by learning to rotate a cube in-hand.
Their system can successfully execute rotations to target faces in a real
robotic setup despite having learned only in simulation.

\begin{figure}
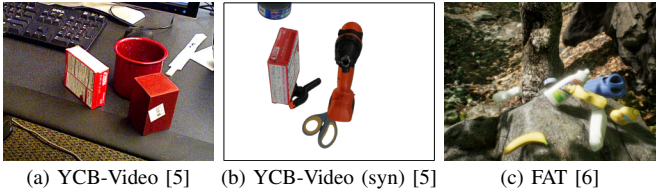

 \centering
 \newlength{\relw}\setlength{\relw}{2.8cm}
 \setlength{\tabcolsep}{0.05cm}\footnotesize
 \begin{tabular}{ccc}
 \includegraphics[width=\relw]{images/related/ycb.png} &
 \includegraphics[frame,width=\relw]{images/related/ycb_syn.png} &
 \includegraphics[width=\relw,clip,trim=160 0 160 60]{images/related/fat.jpg} \\
 (a) YCB-Video \citep{xiang2018posecnn} &
 (b) YCB-Video (syn) \citep{xiang2018posecnn} &
 (c) FAT \citep{tremblay2018falling}
 \end{tabular}
 \caption{YCB-Video dataset and related synthetic datasets.}
 \label{fig:related}
\end{figure}

\begin{figure*}
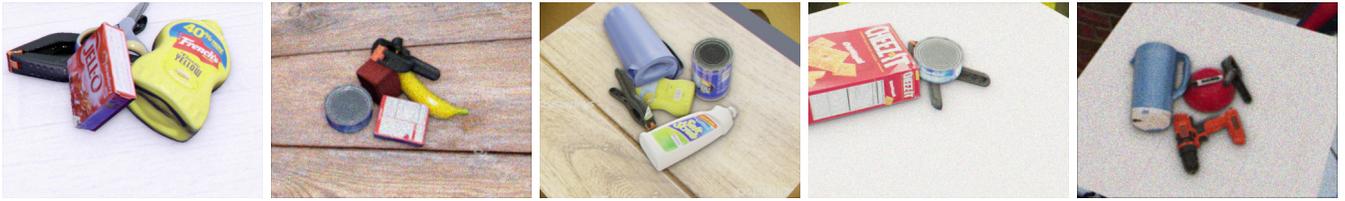

 \centering
 \newlength{\genimgw}\setlength{\genimgw}{3.45cm}
 \includegraphics[width=\genimgw]{images/generator1/000001.png}
 \includegraphics[width=\genimgw]{images/generator1/000003.png}
 \includegraphics[width=\genimgw]{images/generator1/000006.png}
 \includegraphics[width=\genimgw]{images/generator1/000007.png}
 \includegraphics[width=\genimgw]{images/generator1/000010.png}
 \caption{Scenes generated by our physics-based arrangement engine.
 Object meshes are taken from the YCB Object set. Background images are randomly
 selected from an Internet image search with keywords ``table surface''.}
 \label{fig:arrangements}
\end{figure*}

\begin{figure*}[b]
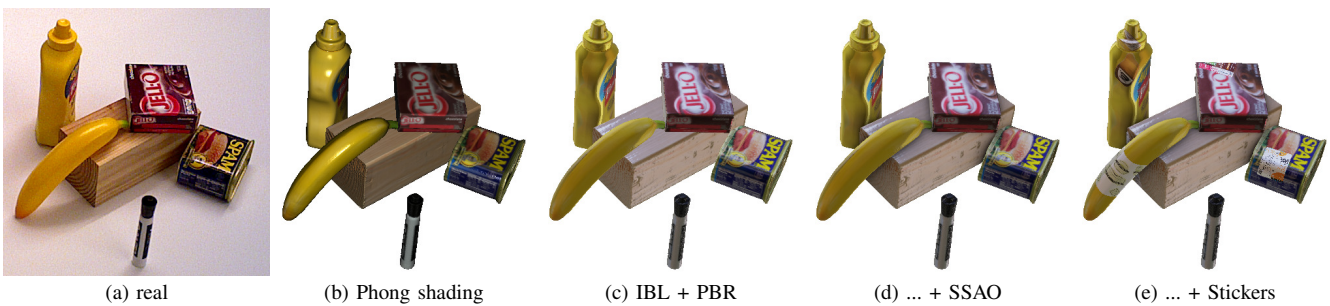

 \centering

 \newlength{\imgw}
 \setlength{\imgw}{3.55cm}
 \setlength{\tabcolsep}{0cm}
 \footnotesize
 \begin{tabular}{ccccc}
  \includegraphics[width=\imgw,clip,trim=200 50 30 20]{images/outputs/sl_real.png} &
  \includegraphics[width=\imgw,clip,trim=200 50 30 20]{images/outputs/sl_phong.png} &
  \includegraphics[width=\imgw,clip,trim=200 50 30 20]{images/outputs/sl_ibl.png} &
  \includegraphics[width=\imgw,clip,trim=200 50 30 20]{images/outputs/sl_ssao.png} &
  \includegraphics[width=\imgw,clip,trim=200 50 30 20]{images/outputs/sl_sticker.png} \\
  (a) real & (b) Phong shading & (c) IBL + PBR & (d) ... + SSAO & (e) ... + Stickers
 \end{tabular}
 \caption{Rendering extensions. For comparison, a real training image from the
 YCB-Video dataset is shown in (a). The other images show the effect of the
 discussed rendering extensions on a synthetic scene rendered with the same
 poses. Note that material settings for PBR (d) are randomized---which is why
 the mustard bottle looks metallic.}
 \label{fig:extensions}
\end{figure*}

Many recent works in pose estimation make usage of synthetic data for training their models.
However, many treat synthetic data as an \textit{augmentation} technique for
real data. For example, \citet{xiang2018posecnn} render synthetic
images in addition to their YCB-Video dataset for training their pose estimation method (see \cref{fig:related}).
\Citet{oberweger2018making} use a similar strategy, but use an intermediate
feature mapping network to learn and abstract away domain differences.
In both works, the use of synthetic data is not a key point and is thus not
systematically analyzed.

\Citet{tremblay2018falling} introduce a large-scale dataset called Falling Things (FAT),
which is rendered from
the YCB Object set, similar to our experiments in \cref{sec:eval}. However,
their method is offline and not focused on dense, cluttered object arrangements.
Furthermore, no comparison of a synthetically trained model against one trained
with real data is attempted.
In our experiments, we compare against a model trained on FAT.
In further work, \citet{tremblay2018deep} present a pose estimation method especially suited
for training on synthetic data. The authors demonstrate that the system can
be successfully trained on a combination of simple domain-randomized images
and the high-quality FAT images and outperforms PoseCNN \citep{xiang2018posecnn},
which was trained on the real YCB-Video Dataset. Whether the increase in performance
is due to the training on synthetic data or due to algorithmic differences to
PoseCNN remains unclear, though.

\Citet{zhang2017physically} present a large-scale synthetic dataset for indoor
scene segmentation. The renders are of very high quality and yield improvements
in segmentation, normal estimation, and boundary detection when used for pretraining.
In contrast to our work, the scenes were manually designed and annotated with
physically realistic material properties. To achieve photorealism, the authors
used a ray tracer for rendering, whereas our OpenGL-based renderer is fast enough
for online usage.

\Citet{kar2019metasim} (published after submission of our work) learn a
generative model for synthesizing annotated 3D scenes. Their approach automatically
tunes many of the hyperparameters introduced in our method, but only optimizes
the scene graph, excluding camera noise modeling and rendering options.
Furthermore, the approach requires a comparison dataset large enough to
estimate the scene distribution, though labels are not needed.

There are other works on online rendering in deep learning settings, usually
for render-and-compare and mesh reconstruction applications. \Citet{kato2018neural}
reconstruct meshes from 2D images using a differentiable rendering module.
Their approach, however, ignores texture.
\Citet{li2018deepim} implement an iterative render-and-compare method for pose
estimation. In contrast to our work, they focus on single objects. Furthermore,
the rendering does not need to be particularly realistic, since a network trained
with pairs of synthetic and real images computes the pose delta for the next iteration.

\section{Method}

Our method design is driven by several goals: To be usable in life-long learning
scenarios (such as the one modeled in the ARC competition), it should be usable
\textit{online}, i.e. without a separate lengthy rendering process. This implies
that the system needs to be \textit{efficient}, especially since we do not want
to take GPU power away from the actual training computations.
It also needs to be \textit{flexible}, so that users can quickly adapt it to
their target domain.

An overview of the architecture is shown in \cref{fig:architecture}.

\subsection{Object Mesh Database}

Input to our pipeline are object meshes. These can come from a variety of
sources: Ideally, 3D scans of the target objects are available, as these
are usually the most precise description of the object geometry and its appearance.
Another possibility are CAD designs. There are also large-scale databases
of 3D meshes which can be applicable, such as ShapeNet~\cite{chang2015shapenet},
although they often lack texture.
Finally, there are online communities such as Sketchfab~\footnote{Sketchfab: \url{https://www.sketchfab.com}},
which allow the retrieval (and purchase) of high-quality mesh models.
For physics simulation in the next pipeline step, the meshes need to be annotated
with inertial properties. If there is no further information, we assume a uniform
density of \SI{500}{\kilogram\per\meter\cubed} and compute mass and inertial
properties from this.

\subsection{Arrangement Engine}

The arrangement engine is responsible for generating physically realistic
object configurations in the given scene. Since we focus on objects
resting in cluttered configurations on a planar support surface, we choose a
plane with random normal\footnote{In case $n_p$ is pointing away from the camera, it is flipped.} $n_p$
and support point $p=(0\;0\;d)^T$ in the camera
coordinate system. The chosen objects (in our experiments five at a time) are
then inserted into the scene at random poses above the plane.
Using the PhysX physics engine by NVIDIA\footnote{PhysX: \url{https://developer.nvidia.com/physx-sdk}},
we then simulate the objects' behavior under gravity in direction $-n_p$ along
with a weaker force of \SI{1}{\newton} drawing the objects towards the support point $p$.
The arrangement engine runs entirely on CPU to spare GPU power for rendering
and the training process itself. To make online operation feasible, we reduce
mesh complexity using Quadric Edge Simplification \citep{garland1997surface}
to a target number of faces $F=2000$. Since PhysX (like all major physics engines)
can only simulate convex dynamic bodies, we then compute a convex hull for each
mesh. This obviously excludes arrangements that make use of concavities, but
initial experiments showed that the impact is negligible.
To cover remaining configurations that are not captured by our arrangement
engine, we randomly fall back to a simpler arrangement procedure that simply
samples collision-free object poses (i.e. without simulating gravity).
Resulting arrangements can be seen in \cref{fig:arrangements}.

\subsection{Rendering}

The rendering process is implemented as standard GPU rasterization using the
Magnum OpenGL engine\footnote{Magnum: \url{https://magnum.graphics}}.
A single rendering pass suffices, as both color and semantic information
such as class and instance segmentation and point coordinates can be written
into the output by a custom fragment shader.

We implemented several extensions that model particular effects arising in
densely cluttered scenes:

\textbf{Physically-based Rendering (PBR)} refers to the employment of
   physically-motivated bidirectional reflectance distribution functions (BRDF).
   The usual choice of the Cook-Torrance BRDF~\citep{cook1982reflectance} has
   two material properties, usually called metalness and roughness.
   Since we assume no prior knowledge, we set these parameters randomly for
   each object on each render.

\textbf{Image-based Lighting (IBL)} allows us to place the object into complex
   lighting situations without explicitly modeling all light sources. This
   yields diffuse and specular light responses on the object surfaces, which
   would be difficult to generate explicitly.

\textbf{Ambient Occlusion (SSAO)} is an approximation for the dark shadows caused by
   close arrangements of objects. Since the light causing these shadows is
   ambient, i.e. caused by multiple light bounces of environment and object
   surfaces, traditional shadow mapping techniques cannot model this effect
   within the rasterization framework. Screen Space Ambient Occlusion
   approximates it by sampling possible occluders in screen space.

For more information about these well-known techniques, we refer to the book by \citet{fernando2004gpu}.

\begin{figure*}
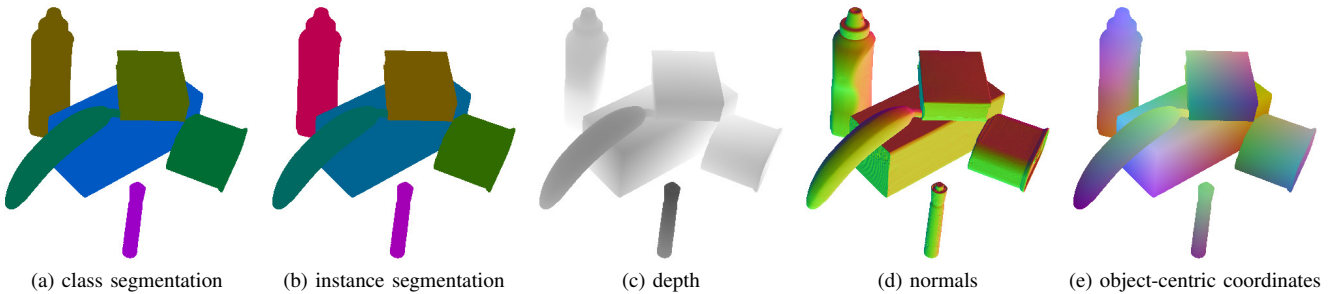

 \centering

 \setlength{\tabcolsep}{0cm}
 \setlength{\imgw}{3.55cm}
 \footnotesize
 \begin{tabular}{ccccc}
  \includegraphics[width=\imgw,clip,trim=200 50 30 20]{images/outputs/sl_classes.png} &
  \includegraphics[width=\imgw,clip,trim=200 50 30 20]{images/outputs/sl_instance.png} &
  \includegraphics[width=\imgw,clip,trim=200 50 30 20]{images/outputs/sl_depth.png} &
  \includegraphics[width=\imgw,clip,trim=200 50 30 20]{images/outputs/sl_normals.png} &
  \includegraphics[width=\imgw,clip,trim=200 50 30 20]{images/outputs/sl_coordinates.png} \\
  (a) class segmentation & (b) instance segmentation & (c) depth & (d) normals & (e) object-centric coordinates
 \end{tabular}
 \caption{Output channels. The library generates a number of different channels
 usable for training of various perception tasks. The scene is the same as shown
 in \cref{fig:extensions}.}
 \label{fig:output_channels}
\end{figure*}

We further developed a texture modification method driven by the fact that
objects are often slightly modified from their original appearance, e.g. by
placing a bar code sticker or product label on them. We model these effects by
randomly projecting an image randomly selected from an Internet search using
the keyword ``product label'' onto the object surface. We call this extension
\textbf{Sticker Projection (SP)}.

\Cref{fig:extensions} shows the effect of the individual rendering extensions.
The contribution of each extension is analyzed in \cref{sec:eval}.

Finally, the background can be customized as well. We render the scene on top
of background images, which can be sampled from image datasets or provided by
the user, for example when training for a well-known structured background environment
in industry applications. Furthermore, the supporting surface can also be textured (see \cref{fig:arrangements}).

The rendering output contains (see \cref{fig:output_channels}):
\begin{itemize}
 \item Pixel-wise color (the image),
 \item pixel-wise class and instance segmentation for training semantic segmentation,
 \item pixel-wise depth and normals for training RGB-D models,
 \item pixel-wise coordinates in each object's coordinate frame for correspondence
   training as in \citep{schmidt2016self}, and
 \item each object's pose.
\end{itemize}

We note that the renderer is equipped with approximative differentiation, allowing
backpropagation of image-space gradients to object pose gradients. This functionality
has been described by \citet{periyasamy2019refining} in detail.

\subsection{Camera Model}

In robotic applications, sensors are often low-cost commodity types, resulting
in imperfect captures of the real scene. Additionally, lighting conditions are
often difficult, yielding even higher noise levels.
Any robust perception model will need to deal with these effects. In the
standard case of training on a larger dataset, the noise statistics can be
learned from the dataset. In our case, we model camera effects with broadly
randomized parameters to obtain a model robust to the perturbations.

We follow the work of \citet{carlson2018modeling}, who propose a comprehensive
camera model including chromatic aberration, blur, exposure, noise, and color
temperature. The operations were implemented in PyTorch using CUDA, so that
rendered images do not have to be copied back to CPU.

\section{Experiments}
\label{sec:eval}

Our experiments are carried out on the YCB-Video dataset~\citep{xiang2018posecnn}.
This dataset is intended for evaluating 6D pose estimation and scene segmentation
methods and contains video sequences captured by a hand-held camera of static
object arrangements. The cluttered arrangements and bad lighting conditions make it highly
challenging. One highly interesting property is that the objects are drawn from
the YCB Object and Model set~\citep{calli2015benchmarking}, for which high-quality
3D meshes are available. We sample random background images from ObjectNet3D~\citep{xiang2016objectnet3d} and supporting plane textures from a top-50 Internet image search with keywords
``table surface''.

We perform all our training experiments using the light-weight RefineNet
architecture~\citep{nekrasov2018light} on the task of semantic segmentation.
Contrary to more high-level tasks, semantic segmentation
is straightforward to evaluate due to its pixel-wise nature and is, in many
current pipelines, a prerequisite for pose estimation in cluttered scenes.

We train all networks using the Adam optimizer with a learning rate of $1e-5$
and parameters $\beta_1=0.9$, $\beta_2=0.999$ for 450k frames. The networks are
evaluated using the mean Intersection-over-Union (IoU) score on the YCB-Video test
set (keyframes).

\subsection{Timings}

We perform our timing tests on a compute server with 2x Intel Xeon Gold 6248 CPUs
running at \SI{2.5}{\giga\hertz} to \SI{3.9}{\giga\hertz}. For rendering and
training, an NVIDIA TITAN RTX GPU is used. Running stand-alone, our library
generates 640$\times$480 annotated training images with 30\,fps when using eight arrangement
engine processes in parallel.
When training a RefineNet network using the generated frames, the entire pipeline
achieves 10\,fps using a batch size of four. In comparison, a real dataset
read from disk yields 13\,fps. If the small drop in performance needs to be avoided,
a second GPU could be used for rendering, hiding the rendering cost by rendering
in parallel to training.

\subsection{Semantic Segmentation}

\begin{table}
 \centering
 \caption{Segmentation results and Ablation Study on YCB-Video}
 \small
 \begin{tabular}{lcc}
  \toprule
  Training data source & Mean IoU & Relative  \\
  \midrule
  Real + stillleben & 0.800 & 1.053 \\
  \textbf{Real dataset}      & \textbf{0.760} & \textbf{1.000} \\
  Real dataset (50\%) & 0.697 & 0.917 \\
  Real dataset (25\%) & 0.597 & 0.786 \\
  \midrule
  with real poses   & 0.691 & 0.909 \\
  \textbf{stillleben (ours)} & \textbf{0.656} & \textbf{0.863} \\
  w/o stickers               & 0.635 & 0.836 \\
  w/o SSAO                   & 0.632 & 0.832 \\
  w/o PBR + IBL              & 0.648 & 0.853 \\
  w/o cam model              & 0.128 & 0.168 \\
  \midrule
  Falling Things \citep{tremblay2018falling} & 0.632 & 0.832 \\
  YCB-Video synthetic \citep{xiang2018posecnn} & 0.300 & 0.395 \\
  \bottomrule
 \end{tabular}
 \label{tab:segmentation}
\end{table}

\begin{figure}
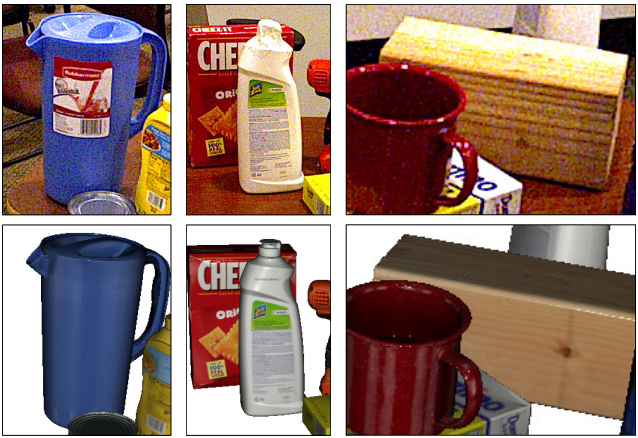

 \centering

 \newlength{\diffimgwidth}
 \setlength{\diffimgwidth}{2.8cm}
 \setlength{\tabcolsep}{0.1cm}
 \begin{tabular}{ccc}
  \includegraphics[frame,height=\diffimgwidth,clip,trim=100 80 300 100]{images/difficult/pitcher/sl_real.png} &
  \includegraphics[frame, height=\diffimgwidth,clip,trim=100 80 300 50]{images/difficult/cleaner/sl_real.png} &
  \includegraphics[frame, height=\diffimgwidth,clip,trim=220 120 200 200]{images/difficult/wood_block/sl_real.png} \\

  \includegraphics[frame, height=\diffimgwidth,clip,trim=100 80 300 100]{images/difficult/pitcher/sl_phong.png} &
  \includegraphics[frame, height=\diffimgwidth,clip,trim=100 80 300 50]{images/difficult/cleaner/sl_phong.png} &
  \includegraphics[frame, height=\diffimgwidth,clip,trim=220 120 200 200]{images/difficult/wood_block/sl_phong.png}
 \end{tabular}
 \caption{Deviations of the real objects from their meshes: The top row shows
 crops of YCB-Video test frames, the bottom row their corresponding renders.
 The pitcher has a sticker on it, the cleaner bottle is wrapped with tape, and
 the wood block seems to have entirely different texture and surface finish.
 Furthermore, the mesh textures exhibit artifacts from the scanning process,
 such as the vertical reflection stripes on the cup.}
 \label{fig:difficult_meshes}
\end{figure}

Models trained with our synthesized data obtain roughly 85\,\% performance compared
to a baseline trained on the real dataset (see \cref{tab:segmentation}).
We anticipated that we would not be
able to outperform the real baseline, since there are many complex effects
our arrangement engine and renderer cannot capture. Additionally, some of the
objects have radical differences to their object meshes, some of which our
sticker projection module cannot produce (see \cref{fig:difficult_meshes}).
To investigate the remaining gap, we also performed a training run with real
poses, i.e. object configurations taken from the YCB-Video training set, but
rendered using our pipeline. As this model achieves 91\,\% compared to the baseline, we conclude
that there are both arrangement and visual effects our pipeline cannot yet
capture.

When the size of the real YCB-Video training dataset is artificially reduced,
baseline performance drops below our model trained with synthetic data (see \cref{tab:segmentation}).
As the YCB-Video dataset is quite large (265\,GiB), we conclude that our
synthesis pipeline can replace capture of a large-scale dataset while
attaining comparable performance.

To judge the importance of the implemented rendering extensions, we performed
an ablation study (see \cref{tab:segmentation}). While the camera model with
its blurring and introduction of noise is apparently crucial for bridging the
gap between rendered and real images (we hypothesize this step hides obvious
rendering artifacts which would otherwise be learned), the other techniques
yield more modest improvements.

Finally, we compare the usability of our generated data against two
offline-generated datasets using the YCB-Video objects. The original synthetic
images distributed with the YCB-Video dataset \citep{xiang2018posecnn} yield
very suboptimal performance---but they were never intended for standalone
usage without the real dataset.
We outperform the FAT dataset \citep{tremblay2018falling} by a small margin,
which we think is due to our specialization on dense, cluttered arrangements,
whereas FAT contains more spread-out arrangements with less occlusions.

\subsection{Category-Level Generalization}

\begin{figure}
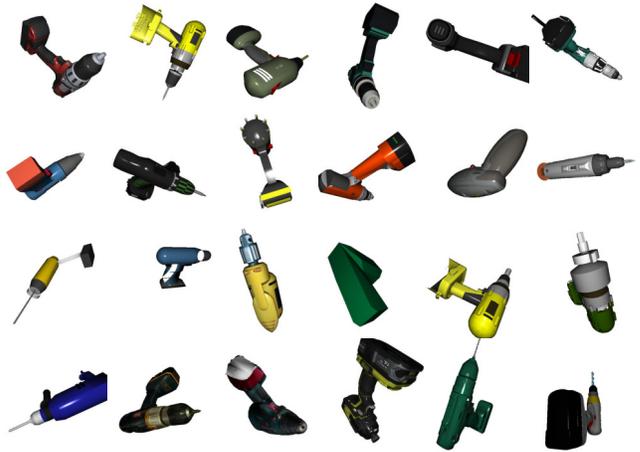

 \centering

 \newlength{\drillw}\setlength{\drillw}{1.4cm}
 \includegraphics[width=\drillw,clip,trim=100 0 100 0]{images/drills/0000.jpg}%
 \includegraphics[width=\drillw,clip,trim=100 0 100 0]{images/drills/0001.jpg}%
 \includegraphics[width=\drillw,clip,trim=100 0 100 0]{images/drills/0002.jpg}%
 \includegraphics[width=\drillw,clip,trim=100 0 100 0]{images/drills/0003.jpg}%
 \includegraphics[width=\drillw,clip,trim=100 0 100 0]{images/drills/0004.jpg}%
 \includegraphics[width=\drillw,clip,trim=100 0 100 0]{images/drills/0005.jpg}
 \includegraphics[width=\drillw,clip,trim=100 0 100 0]{images/drills/0006.jpg}%
 \includegraphics[width=\drillw,clip,trim=100 0 100 0]{images/drills/0008.jpg}%
 \includegraphics[width=\drillw,clip,trim=100 0 100 0]{images/drills/0009.jpg}%
 \includegraphics[width=\drillw,clip,trim=100 0 100 0]{images/drills/0010.jpg}%
 \includegraphics[width=\drillw,clip,trim=100 0 100 0]{images/drills/0011.jpg}%
 \includegraphics[width=\drillw,clip,trim=100 0 100 0]{images/drills/0012.jpg}
 \includegraphics[width=\drillw,clip,trim=100 0 100 0]{images/drills/0013.jpg}%
 \includegraphics[width=\drillw,clip,trim=100 0 100 0]{images/drills/0015.jpg}%
 \includegraphics[width=\drillw,clip,trim=100 0 100 0]{images/drills/0016.jpg}%
 \includegraphics[width=\drillw,clip,trim=100 0 100 0]{images/drills/0018.jpg}%
 \includegraphics[width=\drillw,clip,trim=100 0 100 0]{images/drills/0019.jpg}%
 \includegraphics[width=\drillw,clip,trim=100 0 100 0]{images/drills/0020.jpg}
 \includegraphics[width=\drillw,clip,trim=100 0 100 0]{images/drills/0021.jpg}%
 \includegraphics[width=\drillw,clip,trim=100 0 100 0]{images/drills/0022.jpg}%
 \includegraphics[width=\drillw,clip,trim=100 0 100 0]{images/drills/0023.jpg}%
 \includegraphics[width=\drillw,clip,trim=100 0 100 0]{images/drills/0024.jpg}%
 \includegraphics[width=\drillw,clip,trim=100 0 100 0]{images/drills/0025.jpg}%
 \includegraphics[width=\drillw,clip,trim=100 0 100 0]{images/drills/0026.jpg}%
 \caption{Collection of 24 driller meshes for category-level generalization
 experiment. Meshes obtained from \url{https://sketchfab.com}.}
 \label{fig:driller}
\end{figure}

\begin{figure}
 \begin{tikzpicture}[font=\footnotesize]
  \begin{axis}[ymin=0.3,ymax=1,xmin=-0.2,xmax=100.5,
     xlabel={No. of real images used},
     ylabel={IoU (driller class)},
     height=4cm,
     width=\linewidth,
     ]
    \addplot+ coordinates {
      (0, 0.3782651424407959)
      (5, 0.4079049825668335)
      (10, 0.43253761529922485)
      (25, 0.5649039149284363)
      (50, 0.6099928021430969)
      (100, 0.7495308518409729)
    };

    \addplot+[mark=none, green, samples=2, domain=-10:200] {0.824512243270874};
  \end{axis}
 \end{tikzpicture}\vspace{-0.4cm}
 \caption{Category-level generalization and improvement using few real images.
 We show the segmentation performance
 (specifically the driller class) vs. the number of real images from YCB-Video
 that were used during training. The green line shows the IoU score of the
 model trained on the full YCB-Video training set.}
 \label{fig:generalization}
\end{figure}
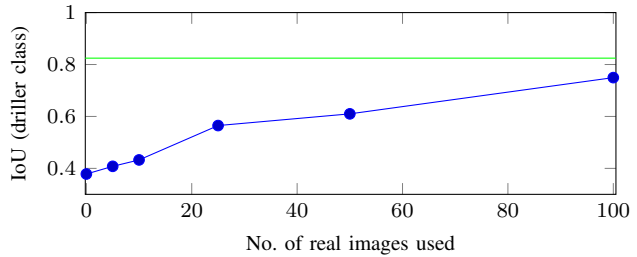

The YCB-Video Dataset recommends itself to these experiments, since it contains
high-quality meshes corresponding to the exact instances used in the dataset.
We maintain that this is a valid use case: Commercial 3D scanners can generate high-quality
3D models from real objects in minutes, which can then be used to train
perception systems using our pipeline.
However, we are also interested in the case where meshes of the exact instance
are not available.
To investigate the usability of our generated data to train class-level
perception that generalizes to unseen instances, we perform an additional
experiment on YCB-Video. Here, we select a particular object, the driller, since
meshes of drillers are readily available. We withhold the original mesh from
our training pipeline and replace it with a collection of 25 driller meshes
collected from online databases such as Sketchfab (see \cref{fig:driller}).

We show the resulting performance on the YCB-Video test set (focused on the
driller class) in \cref{fig:generalization}.
The resulting IoU score is quite low, probably unusable in real-world applications.
We conclude that our mesh collection is insufficient to allow the network to
generalize to the test instance, maybe due to bad mesh quality or too low variety,
allowing the network to learn the different instances by heart.
In this situation, the user may augment the synthetic training data by a few
manually annotated real images, which will improve accuracy to a usable level (see \cref{fig:generalization}).

\subsection{Application in a real-world robotic system}

While we were not able to test the system using real YCB Objects, it was successfully
used for training a CNN-based semantic segmentation and pose estimation in a functional
regrasping pipeline by \citet{pavlichenko2019regrasping}. While
\citep{pavlichenko2019regrasping} contains the system-level description and
details about the manipulation planning aspects, we describe the scene synthesis
step here.

\begin{figure}
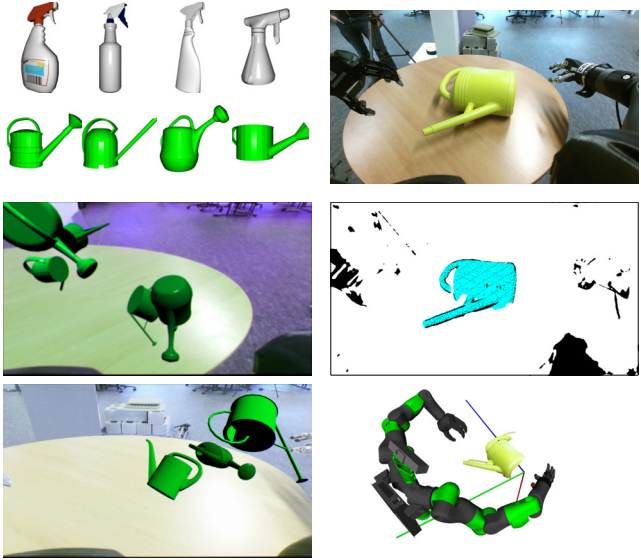

 \centering
 \newlength{\imgwidth}
 \setlength{\imgwidth}{1.2cm}
 \setlength{\tabcolsep}{0.1cm}
 \begin{tabular}{cc}
 \begin{tabular}{c@{}c@{}c@{}c}
  \includegraphics[height=\imgwidth,clip,trim=100 0 30 0]{images/regrasping/spray_bottles/0000.jpg} &
  \includegraphics[height=\imgwidth,clip,trim=100 0 30 0]{images/regrasping/spray_bottles/0002.jpg} &
  \includegraphics[height=\imgwidth,clip,trim=100 0 30 0]{images/regrasping/spray_bottles/0003.jpg} &
  \includegraphics[height=\imgwidth,clip,trim=100 0 30 0]{images/regrasping/spray_bottles/0006.jpg} \\[0cm]
  \includegraphics[height=\imgwidth,clip,trim=100 0 30 60]{images/regrasping/watering_cans/0000.jpg} &
  \includegraphics[height=\imgwidth,clip,trim=100 0 30 60]{images/regrasping/watering_cans/0001.jpg} &
  \includegraphics[height=\imgwidth,clip,trim=100 0 30 60]{images/regrasping/watering_cans/0002.jpg} &
  \includegraphics[height=\imgwidth,clip,trim=100 0 30 60]{images/regrasping/watering_cans/0003.jpg}
 \end{tabular}

 & \adjustbox{valign=m}{\includegraphics[height=2.3cm]{images/regrasping/input.png}} \\

 \includegraphics[height=2.3cm]{images/regrasping/train_watering_can/0000.png} &
 \includegraphics[frame,height=2.3cm]{images/regrasping/segmentation_inv.png} \\

 \includegraphics[height=2.3cm]{images/regrasping/train_watering_can/0003.png} &
 \includegraphics[height=2.3cm]{images/regrasping/pose.png}
 \end{tabular}
 \caption{Regrasping application.
 Left column: Mesh database and generated watering can training scenes.
 Right column: Real input image, semantic segmentation (black: positives,
 cyan: selected region with valid depth), estimated 6D pose.
 Note that the segmentation has false positives on the robot arms, since they
 were not part of the training. These were filtered out using kinematic
 information.
 Left part taken from \citep{pavlichenko2019regrasping}.}
 \label{fig:regrasping:meshes}
\end{figure}

The target objects investigated  in this application were watering cans and spray
bottles. Since a 3D scanner was not available, we retrieved 3D models from
Sketchfab and other online databases (see \cref{fig:regrasping:meshes}).
Nearly all meshes were textureless, so we assigned random uniform colors for
the spray bottles and uniform green color to the watering cans.

As background scenes, real images captured by the robot, observing an empty scene
under different lighting conditions were used. Since the scenes were not expected
to be cluttered, only the basic arrangement engine mode was used, where objects
are placed randomly in a collision-free manner.

Segmentation was performed using the RefineNet architecture (as before), which
was extended to also densely estimate the direction to the object center and
the object's rotation as a Quaternion. Combined, this can be used to estimate
the full 6D pose \citep{xiang2018posecnn}.
The trained model was used successfully in 53 manipulation trials, out of which
the object was segmented with a success rate of 100\%. The subsequent highly
complex manipulation task was completed with a success rate of 65\% \citep{pavlichenko2019regrasping}.
In this setting, which is less complicated than the highly cluttered YCB-Video
scenes, our pipeline resulted in robust category-level segmentation and
pose estimation without any data capture or manual annotation.

\section{Discussion \& Conclusion}

We have presented a pipeline for generating synthetic scenes of densely
cluttered objects from object meshes.
Our generated data can be used to train deep segmentation models to comparable
performance on the challenging YCB-Video dataset. Additionally, we demonstrated
a robotic application including segmentation and pose estimation.
We conclude that in these situations, the time- and labor-consuming task of
capturing and annotating real datasets can be minimized or skipped altogether
in favor of synthetically generated scenes.
The pipeline is fast enough for online usage, which opens up new and exciting
research opportunities, both in online learning systems that can quickly adapt
to changing objects and environments and in iterative systems that can create
and compare internal representations to their sensor input by rendering.

We explicitly note that the pipeline is easy to adapt to different object sets
and environments---and can thus hopefully lower the barrier to training robust
perception systems for custom robotic applications.

Limitations include the dependence on high-quality mesh input, as
demonstrated by our generalization experiments. While remaining gaps between
training set and real objects can be
bridged by the means of additionally captured real images, a purely synthetic
solution would be preferable.
Furthermore, our arrangement procedure leads to entirely random configurations,
whereas humans often place objects in functional ways. To capture this bias,
more complex arrangement options could be investigated.

\section*{Acknowledgment}

\noindent{\footnotesize This research has been supported by an Amazon Research
Award.}

\printbibliography

\end{document}